\documentclass[10pt,twocolumn,letterpaper]{article}

\usepackage[pagenumbers]{cvpr} 

\usepackage{graphicx}
\usepackage{amsmath}
\usepackage{amssymb}
\usepackage{bm}
\usepackage{booktabs}
\usepackage{multirow, makecell}
\usepackage{algorithm}
\usepackage{algorithmic}
\usepackage[accsupp]{axessibility}

\usepackage[pagebackref,breaklinks,colorlinks]{hyperref}

\usepackage[capitalize]{cleveref}
\crefname{section}{Sec.}{Secs.}
\Crefname{section}{Section}{Sections}
\Crefname{table}{Table}{Tables}
\crefname{table}{Tab.}{Tabs.}

\def\cvprPaperID{5597} 
\def\confName{CVPR}
\def\confYear{2022}

\begin{document}

\title{CPPF: Towards Robust Category-Level 9D Pose Estimation in the Wild}


\author{Yang You, Ruoxi Shi, Weiming Wang\thanks{Cewu Lu and Weiming Wang are the corresponding authors. Cewu Lu is member of Qing Yuan Research Institute and MoE Key Lab of Artificial Intelligence, AI Institute, Shanghai Jiao Tong University, China and Shanghai Qi Zhi institute.}\ , Cewu Lu\footnotemark[1] \\ 
Shanghai Jiao Tong University, China\\
\{qq456cvb, eliphat, wangweiming, lucewu\}@sjtu.edu.cn %
}
\maketitle

\begin{abstract}
   In this paper, we tackle the problem of category-level 9D pose estimation in the wild, given a single RGB-D frame. Using supervised data of real-world 9D poses is tedious and erroneous, and also fails to generalize to unseen scenarios. Besides, category-level pose estimation requires a method to be able to generalize to unseen objects at test time, which is also challenging. Drawing inspirations from traditional point pair features (PPFs), in this paper, we design a novel Category-level PPF (CPPF) voting method to achieve accurate, robust and generalizable 9D pose estimation in the wild. To obtain robust pose estimation, we sample numerous point pairs on an object, and for each pair our model predicts necessary SE(3)-invariant voting statistics on object centers, orientations and scales. A novel coarse-to-fine voting algorithm is proposed to eliminate noisy point pair samples and generate final predictions from the population. To get rid of false positives in the orientation voting process, an auxiliary binary disambiguating classification task is introduced for each sampled point pair. In order to detect objects in the wild, we carefully design our sim-to-real pipeline by training on synthetic point clouds only, unless objects have ambiguous poses in geometry. Under this circumstance, color information is leveraged to disambiguate these poses. Results on standard benchmarks show that our method is on par with current state of the arts with real-world training data. Extensive experiments further show that our method is robust to noise and gives promising results under extremely challenging scenarios. Our code is available on \href{https://github.com/qq456cvb/CPPF}{https://github.com/qq456cvb/CPPF}.
\end{abstract}

\section{Introduction}

\begin{figure}[ht]
    \centering
    \includegraphics[width=0.9\linewidth]{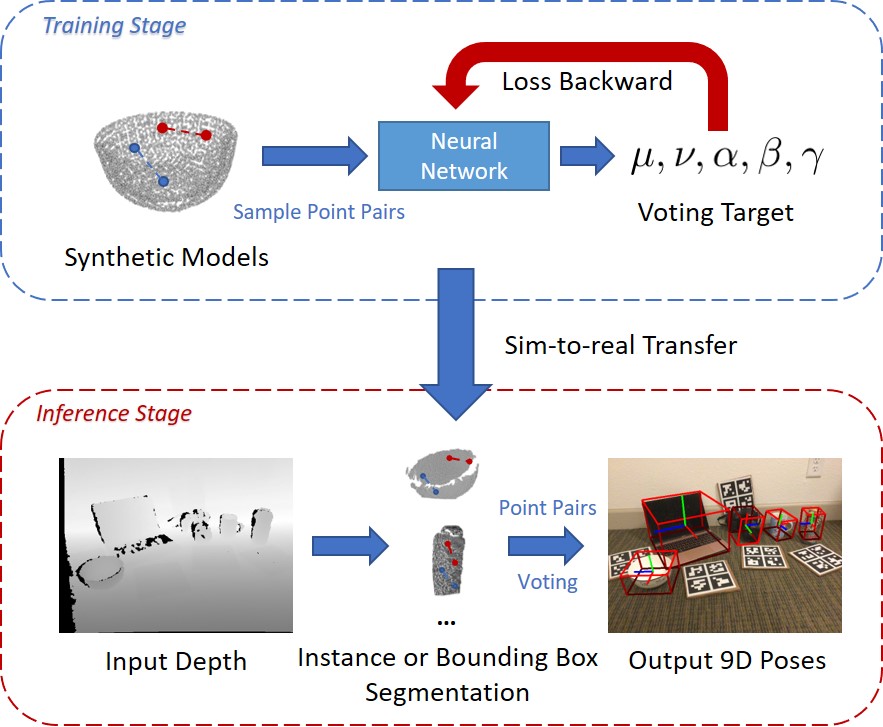}
    \caption{\textbf{An overview of our proposed voting scheme.} During training, for each sampled point pairs on synthetic models, we train a set of voting targets with a \textit{SE(3)} invariant neural network. When testing, objects are first segmented out, and then for each object we randomly sample some point pairs to vote for the final 9D poses (translation, orientation and scale).}
    \label{fig:intro}
\end{figure}

Estimation of 3D position, orientation and scale of novel objects, namely, category-level 9D pose estimation in the wild, is of great importance in many fields, such as robotics~\cite{gao20206d, wang2019normalized,liu2022robotic} and human-object interactions~\cite{ahmadyan2021objectron,li2020detailed}. There are many prior works exploring this direction, but with limitations, though. Some past works~\cite{drost2010model,vidal2018method,brachmann2014learning,rios2013discriminatively,kehl2016deep} have explored the instance-level 6D pose estimation. However, they require exact object models and their sizes beforehand, which is often not realizable in real-world scenarios. NOCS~\cite{wang2019normalized} introduces normalized object coordinate space to give a consistent representation across intra-class objects. Although it is able to achieve category-level pose estimations, it requires real-world pose annotations, which are tedious and limited by size. Besides, the 3D object scales predicted by NOCS are simple heuristics and prone to object occlusions, which is inevitable in real world. We doubt if one can leverage a sim-to-real approach that generalize 9D pose estimations from synthetic objects to real world, since ground-truth pose annotations in real-world scenarios are hard to acquire. Gao \textit{et al.}~\cite{gao20206d} tries to solve this problem via comparing the appearance of objects of synthetic objects and real objects, and finds a pose that minimize the difference. Though this method does not require real 9D pose labels, it is erroneous and inferior to NOCS in terms of orientation error, due to the domain gap between synthetic and real RGB images.

Embracing these challenges, in this paper, we propose Category-level PPF (CPPF) that votes for category-level 9D poses.  We draw some inspirations from traditional point pair features (PPFs), and formulate the problem of pose estimation as a voting process, where each point pair would generate several offsets or relative angles towards ground-truth 9D poses. Next, the pose with the most votes is cast as our final prediction. In contrast to traditional instance-level PPFs, where each pair is matched against an offline database, our method is much faster and able to generalize to unseen objects. To overcome the difficulty in orientation voting, where false positives are generated, an auxiliary binary classification task is introduced. 

In order to segment out the point cloud of a real-world target object, we leverage an off-the-shelf or fine-tuned instance segmentation model. We argue that instance segmentation labels are much cheaper and easier to annotate than 9D poses.

Besides, we develop a two-stage coarse-to-fine voting method, to robustly estimate the object pose when predictions of instance segmentation are not accurate. This method could eliminate noisy point pairs that do not contribute to the voting of object pose. Furthermore, our additional experiments show that when only object bounding boxes are available, our method could still achieve robust and appealing results.

We evaluate our method on the publicly available real-world dataset released by Wang \textit{et al.}~\cite{wang2019normalized} on category-level object pose estimation. Results show that our method beats sim-to-real state of the arts and is comparable to real-world training methods. 
In addition, we show that when only bounding box detections are provided, our method still gives decent 9D pose predictions.
To further evaluate the generalization ability of our method to real-world scenarios, we directly apply our method on SUN RGB-D~\cite{song2015sun} dataset with zero-shot transfer, which contains much more diverse and complex scenes. 
Our method also outperforms baselines by a large margin. In summary, our contribution is:
\begin{itemize}
    \item We propose a novel category-level voting scheme to extract 9D pose of objects. An auxiliary task is introduced to remove the ambiguity in orientation voting. A coarse-to-fine voting algorithm is proposed to eliminate noisy point pairs with robust pose predictions.
    \item We introduce a novel sim-to-real pipeline with carefully designed point pair features to achieve generalizable sim-to-real transfer, with synthetic models only.
    \item Extensive experiments show that our sim-to-real method is on par with current state-of-the-art methods, which utilize real-world training data. Besides, our model is robust to segmentation errors and could give accurate pose predictions with only bounding box detections.
\end{itemize}

\section{Related Works}
\subsection{Object Pose Estimation}
\paragraph{Instance-Level Pose Estimation} There are many pose detection methods that requires only point clouds. Drost \textit{et al.}\cite{drost2010model} propose point pair features to match against an object database using a voting scheme. Later, some researchers improves upon Drost's work: Hinterstoisser \textit{et al.}\cite{hinterstoisser2016going} leverage a smart sampling scheme to restrict the searching range; Vidal \textit{et al.}\cite{vidal2018method} improve the matching process by considering neighborhoods that potentially affected due to noise. It also improves the post-processing step such that the retrieved pose is more consistent with the observed camera view. Shi \textit{et al.}\cite{shi2021stablepose} propose a method on generating object poses from stable geometric groups. There are also many works taking RGB(-D) images as input. Kehl \textit{et al.}\cite{kehl2017ssd} extend the popular SSD paradigm to cover the full 6D pose space. Branchmann \textit{et al.}\cite{brachmann2014learning} learn to classify each pixel into a set of normalized coordinates and then generates a set of candidates by RANSAC. 
Grabner \textit{et al.}\cite{grabner20183d} render depth images from 3D models using the predicted poses and match learned image descriptors of RGB images against those of rendered depth images using a CNN-based multi-view metric learning approach.
Rios \textit{et al.}\cite{rios2013discriminatively} use a discriminative learning approach to match the object pose in images against a database. DeepIM~\cite{li2018deepim} leverages a FlowNet to output relative pose between real and rendered image patches. The pose is refined in an iterative way. 
Kehl \textit{et al.}\cite{kehl2016deep} learn to auto-encode RGB-D patches and match them in a codebook to vote for the final pose. Gao \textit{et al.}\cite{gao20206d} directly regress 6D poses from object point clouds. These methods lack the ability to generalize to unseen objects, and most of them~\cite{drost2010model,vidal2018method,brachmann2014learning,rios2013discriminatively,kehl2016deep} do not scale well.

\paragraph{Category-Level Pose Estimation}
Recently, a few works focus on category-level object estimation, where an unseen object's pose is to be detected. NOCS~\cite{wang2019normalized} learns to regress objects' normalized coordinates establishing 2D-3D relationships, so that object poses can be solved in closed form. However, it requires real-world pose annotations for training. SPD~\cite{tian2020shape} improves the predictions of canonical object models by deforming categorical shape priors. Then, CASS~\cite{chen2020learning} use a variational auto-encoder to capture pose-independent features, along with pose-dependent ones, to directly predict the 6D poses. FS-Net~\cite{chen2021fs} proposes
a decoupled rotation mechanism that uses two decoders to decode the category-level rotation information. For translation and size estimation, it uses a residual estimation network. DualPoseNet~\cite{Lin_2021_ICCV} leverages two parallel decoders either make a pose prediction explicitly, or implicitly do so by reconstructing the input point cloud in its canonical pose. The explicit prediction is then refined with the implicit one. Chen \textit{et al.}\cite{chen2020category} propose to render synthetic models and compare the appearance with real images under different poses. This method, though achieves sim-to-real transfer, is inferior to our method, due to the domain gap between synthetic and real RGB images. Their method, however, is prone to occlusion and noise.

\subsection{Sim-to-Real Transfer}
Sim-to-Real is a common strategy in many fields like object reconstruction, pose estimation and reinforcement learning for robots. ShapeHD~\cite{wu2018learning} and MarrNet~\cite{wu2017marrnet} render realistic ShapeNet~\cite{shapenet2015} models for 3D object reconstruction from a single RGB image. PoseCNN~\cite{xiang2018posecnn} renders different objects into random background to synthesis images for training of object poses. Many works~\cite{labbe2020cosypose,wang2019normalized,hodan2020epos,hodavn2020bop} follow this paradigm, and training on synthetic RGB images have been a common practice in object pose estimation. In reinforcement learning, domain adaption methods~\cite{james2019sim,bousmalis2018using} are usually leveraged, and visual/physical realistic simulation of real environment~\cite{yan2017sim,todorov2012mujoco} plays an important role. Most existing methods explore the domain transfer in color space, while few works~\cite{uy2019revisiting} focus on the domain gap between synthetic and real point clouds. This is due to the fact that in real scenarios, objects are often occluded with noisy backgrounds. 
\section{Preliminaries: Point Pair Features}
In this section, we briefly discuss the original point pair features (PPFs) proposed by Drost \textit{et al.}~\cite{drost2010model}, which can be leveraged for instance-level retrieval.

Given two points $\mathbf{p}_1$ and $\mathbf{p}_2$ with normals $\mathbf{n}_1$ and $\mathbf{n}_2$, set $\mathbf{d} = \mathbf{p}_2 - \mathbf{p}_1$ and define the so-called point pair features $F$:
\begin{align}
\label{eq:ppf}
    F(\mathbf{p}_1, \mathbf{p}_2) = (\|\mathbf{d}\|_2,\angle(\mathbf{n}_1,\mathbf{d}),\angle(\mathbf{n}_2,\mathbf{d}),\angle(\mathbf{n}_1,\mathbf{n}_2)),
\end{align}
In the offline phase, the global model description is created. Such a global model description contains all the pre-calculated PPFs for the object of interest.

In the online phase, a set of reference points in the scene is selected. All other points in the scene are paired with the reference points to create point pair features. These features are matched to the model features contained in the global model description, and a set of potential matches is retrieved. Every potential match votes for an object pose by using an efficient voting scheme where the pose is parametrized relative to the reference point. Specifically, for each match, the pose can be retrieved by aligning the PPF in the scene to that in the offline database. 
For more details, we refer the reader to Drost \textit{et al.}~\cite{drost2010model}. Though Drost PPF has been successful in many scenarios, it can not do a category-level pose estimation, and it is not scalable as the number of objects goes large.



\section{Methodology}
To address the problem raised by instance-level PPFs, we propose Category-level PPF (CPPF) - a brand-new method for detecting category-level object poses. Compared with Drost PPF, our method is free of an offline database. Instead, we directly predict the necessary statistics to vote for object centers, orientations and scales, using a neural network with augmented point pair features as the input.

We assume the target object is first segmented out by some off-the-shelf or fine-tuned instance segmentation model. The segmentation does not need to be exact, and we will introduce our coarse-to-fine voting process to robustly estimate 9D poses when the segmentation is inaccurate. Furthermore, in Section~\ref{sec:exp_nocs_nomask}, we show that instance segmentation can be replaced by coarse bounding box masks.

\subsection{Point Pair Voting}
\subsubsection{Voting for Centers}
\label{sec:votecenter}
Denote the object center as $\mathbf{o}$, for each point pair $\mathbf{p}_1$ and $\mathbf{p}_2$, we predict the following two offsets:
\begin{align}
    \mu &= \overrightarrow{\mathbf{p}_1\mathbf{o}}\cdot \frac{\overrightarrow{\mathbf{p}_1\mathbf{p}_2}}{\|\overrightarrow{\mathbf{p}_1\mathbf{p}_2}\|_2},\\
    \nu &= \|\mathbf{o} - (\mathbf{p}_1 + \mu\frac{\overrightarrow{\mathbf{p}_1\mathbf{p}_2}}{\|\overrightarrow{\mathbf{p}_1\mathbf{p}_2}\|_2}) \|_2,
\end{align}
Notice that these two offsets are invariant to rotations and translations, because for arbitrary rotation matrix $\mathbf{R}\in SO(3)$ and translation $\mathbf{t}\in \mathbb{R}^3$, the new offsets $\mu'$ and $\nu'$ are:
\begin{align}
    \mu' &= \mathbf{R}\cdot\overrightarrow{\mathbf{p}_1\mathbf{o}}\cdot \frac{\mathbf{R}\cdot\overrightarrow{\mathbf{p}_1\mathbf{p}_2}}{\|\mathbf{R}\cdot\overrightarrow{\mathbf{p}_1\mathbf{p}_2}\|_2} = \mu,\\
    \nu' &= \|(\mathbf{R}\cdot\mathbf{o}+\mathbf{t}) - (\mathbf{R}\cdot\mathbf{p}_1+\mathbf{t} + \mu\frac{\mathbf{R}\cdot\overrightarrow{\mathbf{p}_1\mathbf{p}_2}}{\|\mathbf{R}\cdot\overrightarrow{\mathbf{p}_1\mathbf{p}_2}\|_2})\|_2 = \nu.
\end{align}
The proof is simple and omitted, observing that both inner product and L2 norm are invariant to rotations.

\begin{figure}[ht]
\centering
\begin{subfigure}[b]{.5\linewidth}
  \centering
  \includegraphics[width=.9\linewidth]{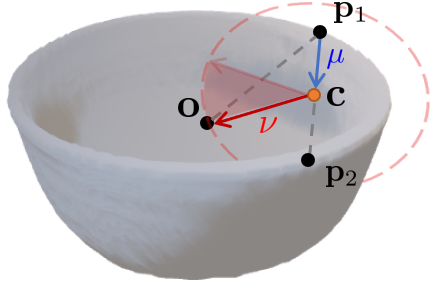}
  \caption{}
  \label{fig:centerstat}
\end{subfigure}%
\begin{subfigure}[b]{.5\linewidth}
  \centering
  \includegraphics[width=\linewidth]{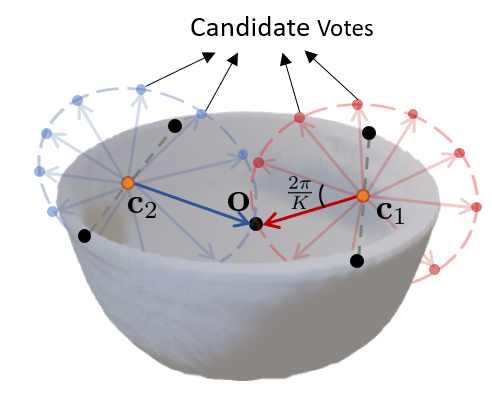}
  \caption{}
  \label{fig:centervotes}
\end{subfigure}
\caption{\textbf{(a) Offset prediction in center voting.} Take a bowl as an example, our model predicts
    $\mu=\|\protect\overrightarrow{\mathbf{p}_1\mathbf{c}}\|_2$ and $\nu=\|\protect\overrightarrow{\mathbf{c}\mathbf{o}}\|_2$, where $\mathbf{c}$
    is the perpendicular foot on $\protect\overrightarrow{\mathbf{p}_1\mathbf{p}_2}$ opposite $\mathbf{o}$. Once $\mu$ and $\nu$ are fixed, object center would possibly lie on the red dash circle. \textbf{(b) Center voting scheme.} For each point pair, candidate centers are generated on the dash circle for an interval of $\frac{2\pi}{K}$. }
\end{figure}

Once $\mu$ and $\nu$ are fixed, the object center is determined up to one degree-of-freedom ambiguity. Specifically, the object center will lie on a circle, with center $\mathbf{c} = \mathbf{p}_1 + \mu\cdot\frac{\overrightarrow{\mathbf{p}_1\mathbf{p}_2}}{\|\overrightarrow{\mathbf{p}_1\mathbf{p}_2}\|_2} $ and radius $\nu$, demonstrated in Figure~\ref{fig:centerstat}.


Inspired by Canonical Voting~\cite{you2022canonical}, we can enumerate every $\frac{2\pi}{K}$ degree and generate multiple votes along the circle. Though there is a circle ambiguity for a single pair, when there are enough pairs, the ground truth location will emerge with the largest vote count, shown in Figure~\ref{fig:centervotes}.



\begin{figure}[ht]
    \centering
    \includegraphics[width=0.4\linewidth]{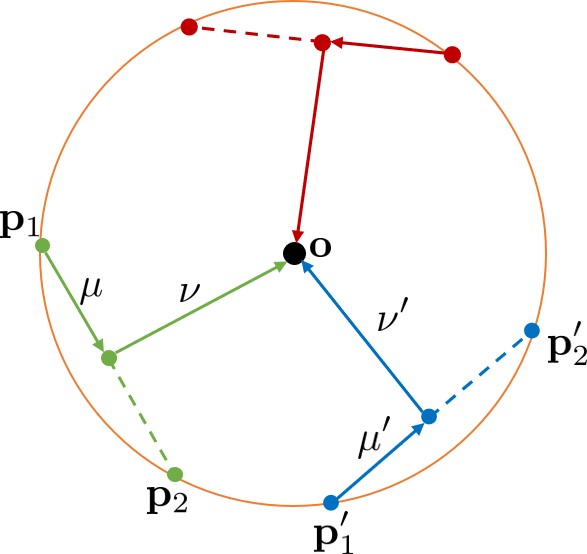}
    \caption{\textbf{Handling symmetric objects.} Here, we take the bird-eye view of a bowl for better illustration. The three symmetric point pairs around the rim share exactly the same input PPF features and output offsets ($\mu$ and $\nu$).}
    \label{fig:symmetry}
\end{figure}

\paragraph{Symmetry in Objects}
Another nice property of our voting scheme is that symmetric objects are naturally handled without any special treatment. Previous methods like NOCS~\cite{chen2020category} require a special treatment of symmetric objects because of the ambiguity of the normalized space. They map the same input features to different outputs due to symmetry. 
In contrast, in our model, input features of symmetric point pairs are exactly the same (due to the $SE(3)$ invariance of PPF), and the output offsets for these pairs are also identical, so that our model learns a proper functional mapping. This is illustrated in Figure ~\ref{fig:symmetry}.

\subsubsection{Voting for Orientations}
Denoting the up orientation as $\mathbf{e}_1$ and right orientation as $\mathbf{e}_2$, we predict the following two relative angles:
\begin{align}
    \alpha &= \mathbf{e}_1\cdot\frac{\overrightarrow{\mathbf{p}_1\mathbf{p}_2}}{\|\overrightarrow{\mathbf{p}_1\mathbf{p}_2}\|_2} \\
    \beta &= \mathbf{e}_2\cdot\frac{\overrightarrow{\mathbf{p}_1\mathbf{p}_2}}{\|\overrightarrow{\mathbf{p}_1\mathbf{p}_2}\|_2}.
\end{align}
These two angles are also invariant to arbitrary rotations.

\begin{figure}
\centering
\begin{subfigure}[b]{.5\linewidth}
  \centering
  \includegraphics[width=.6\linewidth]{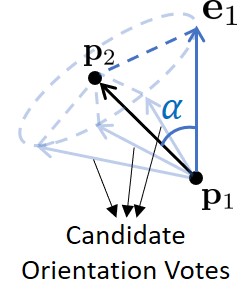}
  \caption{}
  \label{fig:ori1}
\end{subfigure}%
\begin{subfigure}[b]{.5\linewidth}
  \centering
  \includegraphics[width=.9\linewidth]{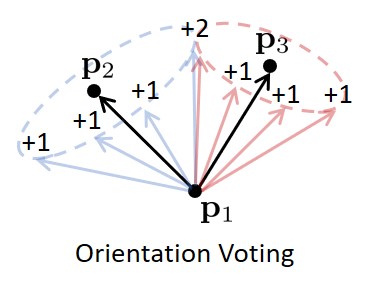}
  \caption{}
  \label{fig:ori2}
\end{subfigure}
\caption{\textbf{(a)} Once $\alpha$ is predicted, the candidate orientation vector will lie on a cone with one degree of freedom. \textbf{(b)} For two point pairs $\mathbf{p}_1$, $\mathbf{p}_2$ and $\mathbf{p}_1$, $\mathbf{p}_3$, we generate candidate votes and count them into bins, the final prediction is the one with the most votes.}
\label{fig:test}
\end{figure}

Analogous to center voting, there is also an ambiguity of one degree of freedom, shown in Figure~\ref{fig:ori1}. Likewise, we also generate a set of proposals for each point pair with a constant degree interval, and then select the predicted orientation as the one with the largest voting count. Because the orientation is continuous, in practice, we uniformly enumerate a set of orientations from unit sphere. For each orientation, we count the number of vote candidates that fall into a fixed solid angle around the orientation. The orientation with the largest count is identified as the final prediction. This is illustrated in Figure~\ref{fig:ori2}.

\paragraph{Removing the Ambiguity of Orientations}
Unfortunately, for orientations, a fake peak with opposite direction sometimes appears when the object is symmetric in structure. When the relative angle $\alpha$ is about $\frac{\pi}{2}$ for a majority of point pairs, the opposite orientation to ground-truth (i.e., -$\mathbf{e}_1$) also receives a lot of point votes, giving a false positive. 

To eliminate false positives, an auxiliary task is introduced. For each point pair $\mathbf{p}_1,\mathbf{p}_2$, we calculate $\mathbf{p}_1$'s normal $\mathbf{n}_1$ (normal ambiguity removed by ensuring $\mathbf{n}_1\cdot\overrightarrow{\mathbf{p}_1\mathbf{p}_2} < 0$). Then we do a binary classification on the following two auxiliary variables:
\begin{align}
    \sigma &= \begin{cases}
      1, & \text{if}\ \mathbf{n_1}\cdot\mathbf{e}_1 > 0 \\
      0, & \text{otherwise}
    \end{cases}, \\
    \tau &= \begin{cases}
      1, & \text{if}\ \mathbf{n_1}\cdot\mathbf{e}_2 > 0 \\
      0, & \text{otherwise}
    \end{cases}.
\end{align}

Taking the up orientation as an example, we show how these auxiliary variables can be used to remove the ambiguity in the opposite direction. During inference, for each point pair, $\hat\sigma$ is predicted by our neural network. Denote the orientation candidate after voting as $\hat{\mathbf{e}}_1$, we calculate two additional statistics $\hat\sigma' = \mathbf{n}_1\cdot\hat{\mathbf{e}}_1$ and $-\hat\sigma' = -\mathbf{n}_1\cdot\hat{\mathbf{e}}_1$ for each pair. Then, $\hat\sigma'$ and $-\hat\sigma'$ are compared with $\hat\sigma$.
If the summation of $\mathrm{CrossEntropy}(\hat\sigma', \hat\sigma)$ from all point pairs is smaller than $\mathrm{CrossEntropy}(-\hat\sigma', \hat\sigma)$. we keep $\hat{\mathbf{e}}_1$; otherwise, we flip the sign of $\hat{\mathbf{e}}_1$ and set $\hat{\mathbf{e}}_1 := -\hat{\mathbf{e}}_1$. We will verify the usefulness of this task in our ablation studies.

\subsubsection{Voting for Scales}
Denoting the category-level average bounding box scales as $\bar{\mathbf{s}}\in \mathbb{R}^3$ and the bounding box scale of a particular instance as $\mathbf{s}\in\mathbb{R}^3$, we predict the following statistic:
\begin{align}
    \bm{\gamma} = \mathrm{log}(\mathbf{s}) - \mathrm{log}(\bar{\mathbf{s}}).
\end{align}
During inference, $\bm{\gamma}$ is first averaged among sampled point pairs, and then the predicted scale can be retrieved as $\hat{\mathbf{s}} = \mathrm{exp}(\bm{\gamma})\ast \bar{\mathbf{s}}$, where $\ast$ is the point-wise product.

\subsubsection{Coarse-to-Fine Voting Process}
In previous sections, we describe the overall voting process for 9D poses (i.e., translation, rotation, scale). However, the generated object pose (especially orientation) might be inaccurate due to noisy points when the instance segmentation is not precise. In order to filter out these noisy points, we propose a coarse-to-fine voting algorithm. Specifically, we first vote for object centers with all the points and then back-trace these votes, keeping only the points that generate enough votes close to the voted object center. Once the noisy points are removed, we vote for the object pose again with the filtered points. A formal description of this algorithm is illustrated in Algorithm~\ref{alg:coarse2fine}.

\begin{algorithm}[ht]
\begin{algorithmic}[1]
    \STATE \textbf{Input}: object point cloud $\{\mathbf{p}_1,\dots,\mathbf{p}_N\}$.
    \STATE \textbf{Output}: translation $\hat{\mathbf{t}}$, orientation $\hat{\mathbf{e}}_1, \hat{\mathbf{e}}_2$ and scale $\hat{\mathbf{s}}$.
    \STATE Initialize empty 3D grids $\mathbf{G}$.
    \FOR{\textbf{each} sampled point pair $\mathbf{p}_1$ and $\mathbf{p}_2$ in point cloud}
        \STATE Use neural network to predict $\mu$ and $\nu$.
        \STATE Generate candidate center votes $\mathbf{o}^{(1)},\dots,\mathbf{o}^{(N)}$.
        \STATE Accumulate candidate votes into $\mathbf{G}$ by discretization.
    \ENDFOR
    \STATE Output $\hat{\mathbf{t}} = \text{argmax}(\mathbf{G})$ as the predicted translation.
    \STATE Initialize point pair pool $\mathbf{P} = \{\}$.
    \FOR{\textbf{each} sampled point pair $\mathbf{p}_1$ and $\mathbf{p}_2$ in point cloud}
        \STATE Use neural network to predict $\mu$ and $\nu$.
        \STATE Generate candidate center votes $\mathbf{o}^{(1)},\dots,\mathbf{o}^{(N)}$.
        \FOR{$i = 1,\dots, N$}
            \IF{$\|\mathbf{o}^{(i)} - \hat{\mathbf{t}}\| < \epsilon$}
                \STATE Add $\mathbf{p}_1$ and $\mathbf{p}_2$ to $\mathbf{P}$.
            \ENDIF
        \ENDFOR
    \ENDFOR
    \STATE Initialize discretized orientation grids $\mathbf{E}_1$ and $\mathbf{E}_2$. \STATE Initialize $\bm{\gamma}_{acc} = 0$.
    \FOR{\textbf{each} sampled point pair $\mathbf{p}_1$ and $\mathbf{p}_2$ in $\mathbf{P}$}
    \STATE Use neural network to predict $\alpha$, $\beta$ and $\bm{\gamma}$.
    \STATE $\bm{\gamma}_{acc} += \bm{\gamma}$.
    \STATE Generate candidate orientation votes $\mathbf{e}_1^{(1)},\dots,\mathbf{e}_1^{(N)}$ and $\mathbf{e}_2^{(1)},\dots,\mathbf{e}_2^{(N)}$ with constant interval.
    \STATE Accumulate candidate votes into $\mathbf{E}_1$ and $\mathbf{E}_2$
    \ENDFOR
    \STATE Output $\hat{\mathbf{s}} = \mathrm{exp}(\frac{\bm{\gamma}_{acc}}{|\mathbf{P}|})\ast \bar{\mathbf{s}}$.
    \STATE Output $\hat{\mathbf{e}}_1 = \text{argmax}(\mathbf{E}_1)$ and $\hat{\mathbf{e}}_2 = \text{argmax}(\mathbf{E}_2)$ as the orientation with the most votes.
    \caption{Coarse-to-Fine Voting Algorithm.}
    \label{alg:coarse2fine} 
\end{algorithmic}
\end{algorithm}

\subsection{Sim-to-Real Transfer}
\subsubsection{Transfer with Depth Maps}
One big advantage of our method is that we only need to train on the synthetic models, and then generalize to real-world scenarios. We achieve this by using the depth map during both training and testing phases, and only \textit{local} point features are leveraged as input. We find that depth or point clouds are much more accurate in sim-to-real generalizations. In contrast, color information is harder to transfer in real-world scenarios, because light conditions are really hard to tune in order to generalize. Color information is only leveraged when there are several ambiguous poses that cannot be distinguished from point clouds.

\paragraph{Rendering Synthetic Models through Realistic Self-Occlusion}
For each category, we choose several synthetic topology-correct models from ShapeNetCore55 similar to that in NOCS~\cite{chen2020category}. Then we use OpenGL~\cite{neider1993opengl} to render each model's depth map from a sampled perspective. All points from the back faces get culled in order to simulate self-occlusion.
Notice that compared with NOCS, our method does not need to choose a random background and paste synthetic models onto it. The only requirement is the synthetic model itself. 

\paragraph{Voxelization and Random Jittering}
Another problem of sim-to-real transfer is the different sampling density in simulated and real scenarios. To mitigate the domain gap, during both training and testing, we voxelize input point clouds with a predefined resolution to obtain a constant sampling density. Besides, we observe that both simulated and real point clouds have a grid artifact. This is due to the rasterization of image pixels. To solve this issue, we randomly jitter the simulated and real point clouds, leading to an improvement on the final results.

\subsubsection{Use Color Information to Disambiguate Poses}
For most categories, our rendered depth images (i.e., point clouds) generalize well to real-world. However, for laptop, there are two ambiguous poses. The laptop lid and keyboard base are hard to discriminate with point clouds only, even for humans. To solve this problem, we train an additional network that takes RGB inputs to segment the lid and keyboard. The training data for this network only contains rendered synthetic laptop images with Blender~\cite{blender}, so that our model is still free of real training images. When testing, we calculate the normal of laptop keyboard by RANSAC plane detection on the predicted segmentation. If the voting result is inconsistent with this normal, we replace it with the normal, otherwise the result is unchanged.


\section{Implementation Details}
In practice, we convert the scalar regression problem into a multi-class classification problem by using a list of anchors, and find that this gives us a better result. We use 32
bins for translation and 36 bins for rotation.
In both training and inference stage, we uniformly sample a fixed number of points (20,000 in training, 100,000 in testing) per model/image and predict their corresponding statistics (i.e., $\mu,\nu,\alpha,\beta,\bm{\gamma}$). In the center voting process, the accumulation 3D grid has a resolution of 0.4 cm except for laptop which is 1 cm. The range of 3D grid is the tightest axis-aligned bounding box of input. In the orientation voting process, the orientation grid has a resolution of 1.5 degrees.
\subsection{Network Architecture}
\begin{figure*}[ht]
    \centering
    \includegraphics[width=0.8\linewidth]{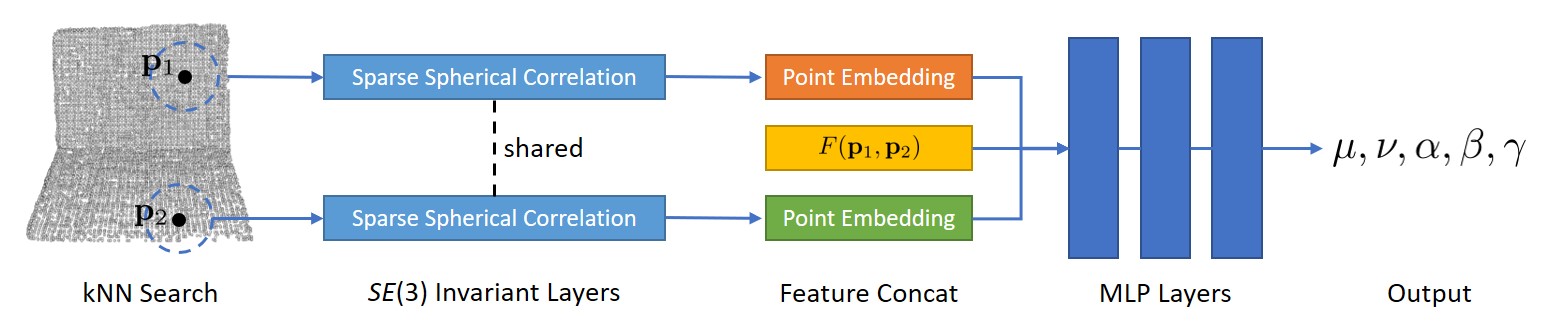}
    \caption{\textbf{Our network architecture.} For each input point pair, we first extract $SE(3)$ invariant embeddings for each point. Then the two embeddings are concatenated with the original PPF feature (Equation~\ref{eq:ppf}), and fed into multi-layer perceptions to predict final outputs. }
    \label{fig:arch}
\end{figure*}

We use SPRIN~\cite{you2021prin}, which is an $SO(3)$ invariant network, and we modified input features to make it $SE(3)$ invariant. The input to the network is the point pair and the set of $k$ nearest neighbors of each point. Denoting $k$ neighbors of point $\mathbf{p}$ as $\{\mathbf{p}^{(1)},\cdots,\mathbf{p}^{(k)}\}$, the sides and angles of triangles formed by $\frac{1}{k}\sum_1^k\mathbf{p}^{(n)}, \mathbf{p}^{(n)}$ and $\mathbf{p}$ are fed to SPRIN to extract rotation invariant point embeddings. Besides, the normal for each point is also estimated and leveraged from its $k$ nearest neighbors. 
The network profession is illustrated in Figure~\ref{fig:arch}. We train each category separately, with Adam optimizer, using learning rate 1e-3, for 200 epochs. The batch size is 1.

\section{Experiments}
In this section, we evaluate our method on two datasets. NOCS REAL275~\cite{wang2019normalized} and SUN RGB-D~\cite{song2015sun}. Both datasets provide RGB-D frames with annotated 9D bounding boxes.
\paragraph{Metrics} We follow NOCS~\cite{wang2019normalized} to report both intersection over union and 6D pose average precision. Intersection over union (IoU) is calculated between the predicted and ground-truth bounding boxes with threshold of 50\%, while 6D pose average precision is calculated by measuring the average precision of object instances for which the error is less than $m$ cm for translation and $n^\circ$ for rotation. We follow NOCS to set a detection threshold of 10\% bounding box overlap between prediction and ground truth to ensure that most objects are included in the
evaluation. Notice that the original 3D box mAP computation code provided by NOCS is buggy. Instead, we use the correct code from Objectron~\cite{ahmadyan2021objectron}.
\subsection{NOCS REAL275 with Instance Mask}
\label{sec:exp_nocs}
Wang \textit{et al.}~\cite{wang2019normalized} captures 8K real RGB-D frames (4300 for training, 950 for validation and 2750 for testing) of 18 different real scenes (7 for training, 5 for validation, and 6 for testing) using a Structure Sensor. We use the 2750 testing scenes for evaluation. We use the instance segmentation masks from NOCS~\cite{wang2019normalized} for a fair comparison.

\paragraph{Comparisons to State of the Arts}
We compare our method with a set of real-world training methods: NOCS~\cite{wang2019normalized}, CASS~\cite{chen2020learning}, SPD~\cite{tian2020shape}, FS-Net~\cite{chen2021fs}, DualPoseNet~\cite{Lin_2021_ICCV}; and a set of methods requiring synthetic training data only: Chen \textit{et al.}~\cite{chen2020category}, Gao \textit{et al.}~\cite{gao20206d}. The original mAP results reported by Chen \textit{et al.}~\cite{gao20206d} uses the IoU matches computed by NOCS~\cite{wang2019normalized} which is potentially unfair, we fix this by using the matches computed by Chen \textit{et al.}'s~\cite{chen2020category} method itself. We also augment Gao \textit{et al.}'s~\cite{gao20206d} method to additionally regress 3D box scales. NOCS~\cite{wang2019normalized}, Chen \textit{et al.}~\cite{chen2020category}, Gao \textit{et al.}~\cite{gao20206d} and DualPoseNet~\cite{Lin_2021_ICCV}'s results are given by running the official code provided by the authors, while the others are borrowed from the original papers. Notice that DualPoseNet~\cite{Lin_2021_ICCV} uses its own instance segmentation masks other than those provided by NOCS, which may result in a higher mAP than the actual.

\begin{figure}[ht]
    \centering
    \includegraphics[width=\linewidth]{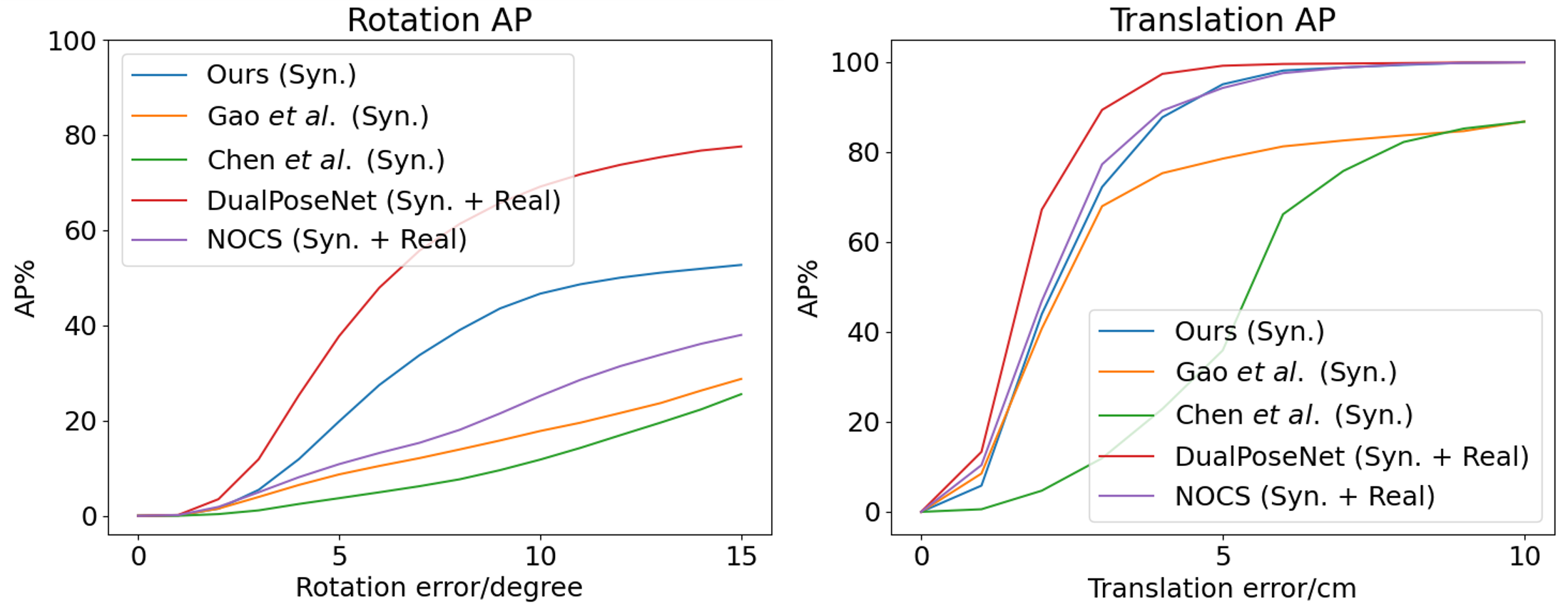}
    \caption{\textbf{Quantitative comparisons on NOCS REAL275 test dataset.} Our method achieves state-of-the-art performance among pure sim-to-real methods.}
    \label{fig:quannocsexp}
\end{figure}
\begin{figure*}[ht]
    \centering
    \includegraphics[width=\linewidth]{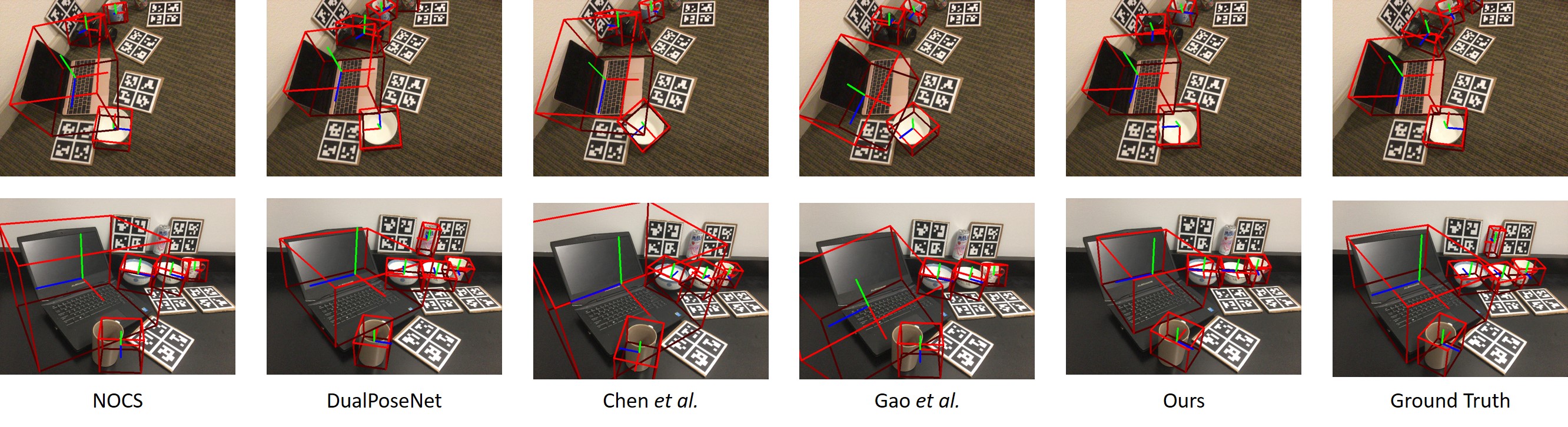}
    \caption{\textbf{Qualitative comparisons on NOCS REAL275 test dataset.}}
    \label{fig:nocsexp}
\end{figure*}

The results are given in Table~\ref{tab:nocs}. We see that our method achieves an mAP of \textbf{16.9}, \textbf{44.9} and \textbf{50.8} for (5$^\circ$, 5 cm), (10$^\circ$, 5 cm) and (15$^\circ$, 5 cm) respectively. It outperforms the best sim-to-real baseline by \textbf{9.1}, \textbf{27.8} and \textbf{24.3}, which is a quite large margin. Our method is also comparable to those methods that are trained on real-world pose annotations. More detailed analysis and comparison is illustrated in Figure~\ref{fig:quannocsexp}. Some qualitative comparisons are given in Figure~\ref{fig:nocsexp}. This experiment shows that our proposed method generalize well to real-world data with only synthetic training data. 

\begin{table}[t]
\begin{center}
\resizebox{\linewidth}{!}{
\begin{tabular}{lcccccc}
\toprule
\multirow{3}*{} & \multirow{3}*{Training Data} & \multicolumn{5}{c}{mAP (\%)}\\
\cmidrule(lr){3-7}
~ & ~ & \multirowcell{2}{3D$_{25}$} & \multirowcell{2}{3D$_{50}$} & \multirowcell{2}{5$^\circ$\\ 5 cm} & \multirowcell{2}{10$^\circ$\\ 5 cm} & \multirowcell{2}{15$^\circ$\\ 5 cm} \\
& & & & & & \\
\midrule
NOCS~\cite{wang2019normalized}  & Syn.(O+B) + Real & 74.4 & 27.8 & 9.8 & 24.1 & 34.9\\
CASS~\cite{chen2020learning} & Syn.(O+B) + Real & - & - & 23.5 & 58.0 & - \\ 
SPD~\cite{tian2020shape} & Syn.(O+B) + Real & - & - & 21.4 & 54.1 & - \\
FS-Net~\cite{chen2021fs} & Syn.(O+B) + Real & - & - & 28.2 & 60.8 & - \\
DualPoseNet~\cite{Lin_2021_ICCV} & Syn.(O) + Real & \textcolor{blue}{82.3} & \textcolor{blue}{57.3} & \textcolor{blue}{36.1} & \textcolor{blue}{67.8} & \textcolor{blue}{76.3} \\
\midrule
Chen \textit{et al.}~\cite{chen2020category} & Syn.(O) & 15.5 & 1.3 & 0.7 & 3.6 & 9.1 \\
Gao \textit{et al.}~\cite{gao20206d} & Syn.(O) & 68.6 & 24.7 & 7.8 & 17.1 & 26.5 \\
\midrule
Ours w/ bbox mask & Syn.(O) & 73.7 & \textcolor{red}{27.2} & 12.4 & 35.3 & 41.2 \\
Ours w/ inst. mask & Syn.(O) & \textcolor{red}{78.2} & 26.4 & \textcolor{red}{16.9} & \textcolor{red}{44.9} & \textcolor{red}{50.8}  \\
\bottomrule
\end{tabular}}
\end{center}
\caption{\textbf{Performance comparison of various methods.} \textit{Syn.(O)} means synthetic ShapeNet objects only; while \textit{Syn.(O+B)} means ShapeNet models rendered with real backgrounds. \textit{Real} means the real-world training data provided by NOCS. The best using real-world training data is marked \textcolor{blue}{blue}, and the best using synthetic training data is marked \textcolor{red}{red}.}
\label{tab:nocs}
\end{table}

\subsection{NOCS REAL275 with Bounding Box Masks}
\label{sec:exp_nocs_nomask}
\begin{figure}[ht]
    \centering
    \includegraphics[width=\linewidth]{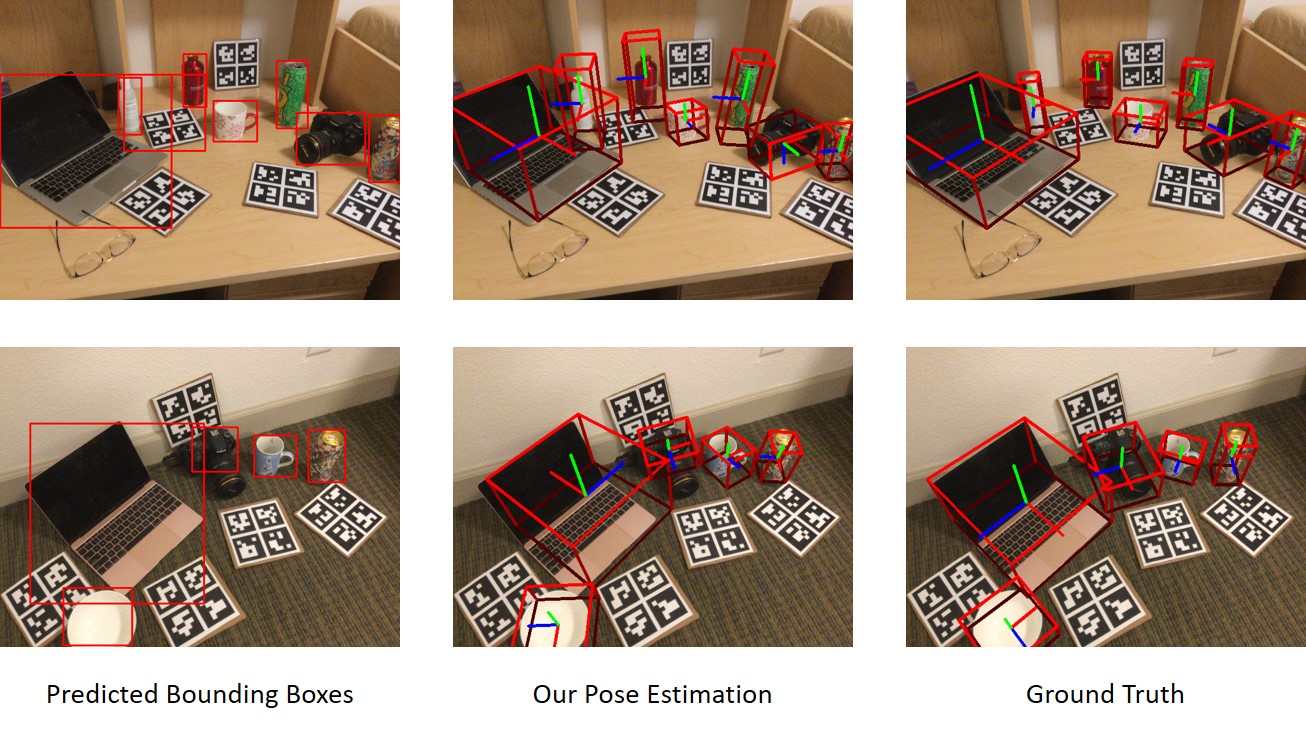}
    \caption{\textbf{Our 9D pose prediction given only bounding box masks on NOCS REAL275 test dataset.}}
    \label{fig:bboxmask}
\end{figure}

Though our method does not require pose annotations for real-world data, it does need to first segment out the target object with a real-world trained instance segmentation model. Can we relax this requirement? Thanks to our robust coarse-to-fine voting algorithm~\ref{alg:coarse2fine} which filters out noisy points, we find that when only bounding boxes are given, our method still achieves significant results. Notice that current state of the arts all require pixel-wise instance segmentation as input. 
Quantitative results are given in Table~\ref{tab:nocs}.
Qualitative pose predictions are shown in Figure~\ref{fig:bboxmask}.

 Moreover, we also tried to get rid of detection priors completely for bowls, more details in the supplementary.
\subsection{SUN RGB-D in the Wild}
\label{sec:wild}
SUN RGB-D~\cite{song2015sun} is a scene understanding benchmark which provides 58,657 9D bounding boxes with accurate object orientations for 10,000 images. We use all the chairs in validation split for evaluation, which contains 2,699 images. In order to make the problem more challenging, we randomly rotate the SUN RGB-D scenes while the original point clouds are aligned with gravity. In addition, we require that all the algorithms cannot see any training data but use existing instance segmentation models (i.e., trained on MSCOCO~~\cite{lin2014microsoft} but not fine-tuned on SUN RGB-D).

\paragraph{Evaluation Results} We compare our method with two baselines: direct back-projection and Gao \textit{et al.}~\cite{gao20206d}. Direct back-projection is a simple baseline that directly back project the detected instance into an axis-aligned bounding box, while Gao \textit{et al.}~\cite{gao20206d} is the same as in Section~\ref{sec:exp_nocs}.
Since this is an extremely difficult task, we only evaluate orientation errors along the up axis. Results are listed in Table~\ref{tab:sunrgbdexp}.
More results are given in the supplementary.

\begin{table}[t]
\begin{center}
\resizebox{0.9\linewidth}{!}{
\begin{tabular}{lccccc}
\toprule
\multirow{3}*{}  & \multicolumn{5}{c}{mAP (\%)}\\
\cmidrule(lr){2-6}
~ & \multirowcell{2}{3D$_{10}$} & \multirowcell{2}{3D$_{25}$} & \multirowcell{2}{20$^\circ$\\ 10 cm} & \multirowcell{2}{40$^\circ$\\ 20 cm} & \multirowcell{2}{60$^\circ$\\ 30 cm} \\
&&&&&\\
\midrule
Back-projection  & 20.2 & 4.4 & 0.0 & 0.6 & 5.8\\
Gao \textit{et al.}~\cite{gao20206d} & 22.2 & 6.0 & 0.0 & 1.0 & 7.0 \\
Ours & \textbf{24.7} & \textbf{8.3} & \textbf{0.8} & \textbf{10.7} & \textbf{17.6}  \\
\bottomrule
\end{tabular}}
\end{center}   
\caption{\textbf{mAP results on SUN RGB-D datasets in the wild.} Our method achieves the best performance.}
\label{tab:sunrgbdexp}
\end{table}

\subsection{Ablation Study and Running Time}
In this section, we conduct various ablation studies on our model. Results are reported on REAL275 test set.
\paragraph{Number of Point Pair Samples and Size of Discrete Orientation Bins}
As the number of pair samples increase, the voting results for orientation and translation become more accurate, while getting saturated for 100,000 point pairs. The size of discrete orientation bins also decides the accumulation accuracy during the voting process. Quantitative results are given in Table~\ref{tab:ablation1}.

\begin{table}[t]
\begin{center}
\resizebox{1.0\linewidth}{!}{
\begin{tabular}{ccccccc}
\toprule
\multirowcell{3}{Number of\\ Point Pairs} & \multirowcell{3}{Orientation\\ Bin Size ($^\circ$)} & \multicolumn{5}{c}{mAP (\%)}\\
\cmidrule(lr){3-7}
 &  & \multirowcell{2}{3D$_{25}$} & \multirowcell{2}{3D$_{50}$} & \multirowcell{2}{5$^\circ$\\ 5 cm} & \multirowcell{2}{10$^\circ$\\ 5 cm} & \multirowcell{2}{15$^\circ$\\ 5 cm} \\
 &&&&&&\\
\midrule
100,000 & 1 & 78.3 & 24.3 & 13.2 & 38.2 & 45.7\\
100,000 & 2 & \textbf{78.5} & 25.9 & 13.4 & \textbf{46.6 }& 52.2 \\
100,000 & 4 & 78.4 & 26.2 & 7.0 & 42.7 & \textbf{53.5} \\
\midrule
10,000 & 1.5 & 68.5 & 21.7 & 8.6 & 27.2 & 34.1\\
60,000 & 1.5 & 78.1 & 25.8 & 15.6 & 42.8 & 49.1 \\
200,000 & 1.5 & 78.4 & \textbf{26.6} & \textbf{17.3} & 45.5 & 51.3\\
\midrule
100,000 & 1.5 & 78.2 & 26.4 & 16.9 & 44.9 & 50.8  \\
\bottomrule
\end{tabular}}
\end{center}   
\caption{\textbf{Effects of sample number and orientation bin size.}}
\label{tab:ablation1}
\end{table}
\paragraph{Auxiliary Task for Disambiguating Poses}
The auxiliary classification task helps our model get rid of potentially flipped orientations, and Table~\ref{tab:ablation2} verifies this.
\paragraph{Whether to use Coarse-to-Fine Voting Process}
Recall, in Algorithm~\ref{alg:coarse2fine}, we filter out point pairs that do not contribute to the proposed object center. This makes our model robust to the noisy points from the imperfect instance segmentation. Quantitative results are given in Table~\ref{tab:ablation2}.
\paragraph{Regression vs. Classification}
Direct regression on relevant statistics are worse than classification. This may due to the fact that regression does not constrain the value into a valid range and produces more noisy outputs. Quantitative results are shown in Table~\ref{tab:ablation2}.
\paragraph{Running Time Analysis}
It takes 171\textit{ms}, 229\textit{ms} and
13\textit{ms} per image for the voting of centers, orientations and scales respectively on a single 1080Ti GPU. Our model is efficient thanks to the highly parallelized voting process.

\begin{table}[t]
\begin{center}
\resizebox{\linewidth}{!}{
\begin{tabular}{lccccc}
\toprule
\multirow{3}*{}  & \multicolumn{5}{c}{mAP (\%)}\\
\cmidrule(lr){2-6}
~ & \multirowcell{2}{3D$_{25}$} & \multirowcell{2}{3D$_{50}$} & \multirowcell{2}{5$^\circ$\\ 5 cm} & \multirowcell{2}{10$^\circ$\\ 5 cm} & \multirowcell{2}{15$^\circ$\\ 5 cm} \\
&&&&&\\
\midrule
No Aux. Classification  & 69.2 & 23.9 & 12.0 & 31.7 & 36.8\\
No Coarse-to-Fine Voting  & 75.6 & 22.3 & 9.7 & 27.4 & 32.6  \\
Regression & 73.3 & 14.7 & 1.8 & 12.7 & 24.2 \\
\midrule
Ours (full) & \textbf{78.2} & \textbf{26.4} & \textbf{16.9} & \textbf{44.9} & \textbf{50.8}  \\
\bottomrule
\end{tabular}}
\end{center}   
\caption{\textbf{Ablation results on NOCS REAL275 test set.}}
\label{tab:ablation2}
\end{table}
\section{Conclusion}
In this paper, we propose a category-level voting algorithm to predict 9D poses in the wild. To overcome the difficulty of false positives during the voting step, an auxiliary orientation classification task is introduced. Our model is trained on synthetic objects and generalizes well to real scenes. Results show that our method is superior to previous sim-to-real methods, even with bounding box masks.

\section{Acknowledgements}
This work was supported by the National Key Research and Development Project of China (No. 2021ZD0110700), the National Natural Science Foundation of China under Grant 51975350, Shanghai Municipal Science and Technology Major Project (2021SHZDZX0102), Shanghai Qi Zhi Institute, and SHEITC (2018-RGZN-02046). This work was also supported by the  Shanghai AI development project (2020-RGZN-02006) and ``cross research fund for translational medicine'' of Shanghai Jiao Tong University (zh2018qnb17, zh2018qna37, YG2022ZD018).

{\small
\bibliographystyle{ieee_fullname}
\bibliography{egbib}

\begin{thebibliography}{10}\itemsep=-1pt

\bibitem{ahmadyan2021objectron}
Adel Ahmadyan, Liangkai Zhang, Artsiom Ablavatski, Jianing Wei, and Matthias
  Grundmann.
\newblock Objectron: A large scale dataset of object-centric videos in the wild
  with pose annotations.
\newblock In {\em Proceedings of the IEEE/CVF Conference on Computer Vision and
  Pattern Recognition}, pages 7822--7831, 2021.

\bibitem{bousmalis2018using}
Konstantinos Bousmalis, Alex Irpan, Paul Wohlhart, Yunfei Bai, Matthew Kelcey,
  Mrinal Kalakrishnan, Laura Downs, Julian Ibarz, Peter Pastor, Kurt Konolige,
  et~al.
\newblock Using simulation and domain adaptation to improve efficiency of deep
  robotic grasping.
\newblock In {\em 2018 IEEE international conference on robotics and automation
  (ICRA)}, pages 4243--4250. IEEE, 2018.

\bibitem{brachmann2014learning}
Eric Brachmann, Alexander Krull, Frank Michel, Stefan Gumhold, Jamie Shotton,
  and Carsten Rother.
\newblock Learning 6d object pose estimation using 3d object coordinates.
\newblock In {\em European conference on computer vision}, pages 536--551.
  Springer, 2014.

\bibitem{shapenet2015}
Angel~X. Chang, Thomas Funkhouser, Leonidas Guibas, Pat Hanrahan, Qixing Huang,
  Zimo Li, Silvio Savarese, Manolis Savva, Shuran Song, Hao Su, Jianxiong Xiao,
  Li Yi, and Fisher Yu.
\newblock {ShapeNet: An Information-Rich 3D Model Repository}.
\newblock Technical Report arXiv:1512.03012 [cs.GR], Stanford University ---
  Princeton University --- Toyota Technological Institute at Chicago, 2015.

\bibitem{chen2020learning}
Dengsheng Chen, Jun Li, Zheng Wang, and Kai Xu.
\newblock Learning canonical shape space for category-level 6d object pose and
  size estimation.
\newblock In {\em Proceedings of the IEEE/CVF conference on computer vision and
  pattern recognition}, pages 11973--11982, 2020.

\bibitem{chen2021fs}
Wei Chen, Xi Jia, Hyung~Jin Chang, Jinming Duan, Linlin Shen, and Ales
  Leonardis.
\newblock Fs-net: Fast shape-based network for category-level 6d object pose
  estimation with decoupled rotation mechanism.
\newblock In {\em Proceedings of the IEEE/CVF Conference on Computer Vision and
  Pattern Recognition}, pages 1581--1590, 2021.

\bibitem{chen2020category}
Xu Chen, Zijian Dong, Jie Song, Andreas Geiger, and Otmar Hilliges.
\newblock Category level object pose estimation via neural
  analysis-by-synthesis.
\newblock In {\em European Conference on Computer Vision}, pages 139--156.
  Springer, 2020.

\bibitem{blender}
Blender~Online Community.
\newblock {\em Blender - a 3D modelling and rendering package}.
\newblock Blender Foundation, Stichting Blender Foundation, Amsterdam, 2018.

\bibitem{drost2010model}
Bertram Drost, Markus Ulrich, Nassir Navab, and Slobodan Ilic.
\newblock Model globally, match locally: Efficient and robust 3d object
  recognition.
\newblock In {\em 2010 IEEE computer society conference on computer vision and
  pattern recognition}, pages 998--1005. Ieee, 2010.

\bibitem{gao20206d}
Ge Gao, Mikko Lauri, Yulong Wang, Xiaolin Hu, Jianwei Zhang, and Simone
  Frintrop.
\newblock 6d object pose regression via supervised learning on point clouds.
\newblock In {\em 2020 IEEE International Conference on Robotics and Automation
  (ICRA)}, pages 3643--3649. IEEE, 2020.

\bibitem{grabner20183d}
Alexander Grabner, Peter~M Roth, and Vincent Lepetit.
\newblock 3d pose estimation and 3d model retrieval for objects in the wild.
\newblock In {\em Proceedings of the IEEE Conference on Computer Vision and
  Pattern Recognition}, pages 3022--3031, 2018.

\bibitem{hinterstoisser2016going}
Stefan Hinterstoisser, Vincent Lepetit, Naresh Rajkumar, and Kurt Konolige.
\newblock Going further with point pair features.
\newblock In {\em European conference on computer vision}, pages 834--848.
  Springer, 2016.

\bibitem{hodan2020epos}
Tomas Hodan, Daniel Barath, and Jiri Matas.
\newblock Epos: Estimating 6d pose of objects with symmetries.
\newblock In {\em Proceedings of the IEEE/CVF conference on computer vision and
  pattern recognition}, pages 11703--11712, 2020.

\bibitem{hodavn2020bop}
Tom{\'a}{\v{s}} Hoda{\v{n}}, Martin Sundermeyer, Bertram Drost, Yann Labb{\'e},
  Eric Brachmann, Frank Michel, Carsten Rother, and Ji{\v{r}}{\'\i} Matas.
\newblock Bop challenge 2020 on 6d object localization.
\newblock In {\em European Conference on Computer Vision}, pages 577--594.
  Springer, 2020.

\bibitem{james2019sim}
Stephen James, Paul Wohlhart, Mrinal Kalakrishnan, Dmitry Kalashnikov, Alex
  Irpan, Julian Ibarz, Sergey Levine, Raia Hadsell, and Konstantinos Bousmalis.
\newblock Sim-to-real via sim-to-sim: Data-efficient robotic grasping via
  randomized-to-canonical adaptation networks.
\newblock In {\em Proceedings of the IEEE/CVF Conference on Computer Vision and
  Pattern Recognition}, pages 12627--12637, 2019.

\bibitem{kehl2017ssd}
Wadim Kehl, Fabian Manhardt, Federico Tombari, Slobodan Ilic, and Nassir Navab.
\newblock Ssd-6d: Making rgb-based 3d detection and 6d pose estimation great
  again.
\newblock In {\em Proceedings of the IEEE international conference on computer
  vision}, pages 1521--1529, 2017.

\bibitem{kehl2016deep}
Wadim Kehl, Fausto Milletari, Federico Tombari, Slobodan Ilic, and Nassir
  Navab.
\newblock Deep learning of local rgb-d patches for 3d object detection and 6d
  pose estimation.
\newblock In {\em European conference on computer vision}, pages 205--220.
  Springer, 2016.

\bibitem{labbe2020cosypose}
Yann Labb{\'e}, Justin Carpentier, Mathieu Aubry, and Josef Sivic.
\newblock Cosypose: Consistent multi-view multi-object 6d pose estimation.
\newblock In {\em European Conference on Computer Vision}, pages 574--591.
  Springer, 2020.

\bibitem{li2018deepim}
Yi Li, Gu Wang, Xiangyang Ji, Yu Xiang, and Dieter Fox.
\newblock Deepim: Deep iterative matching for 6d pose estimation.
\newblock In {\em Proceedings of the European Conference on Computer Vision
  (ECCV)}, pages 683--698, 2018.

\bibitem{li2020detailed}
Yong-Lu Li, Xinpeng Liu, Han Lu, Shiyi Wang, Junqi Liu, Jiefeng Li, and Cewu
  Lu.
\newblock Detailed 2d-3d joint representation for human-object interaction.
\newblock In {\em CVPR}, 2020.

\bibitem{Lin_2021_ICCV}
Jiehong Lin, Zewei Wei, Zhihao Li, Songcen Xu, Kui Jia, and Yuanqing Li.
\newblock Dualposenet: Category-level 6d object pose and size estimation using
  dual pose network with refined learning of pose consistency.
\newblock In {\em Proceedings of the IEEE/CVF International Conference on
  Computer Vision (ICCV)}, pages 3560--3569, October 2021.

\bibitem{lin2014microsoft}
Tsung-Yi Lin, Michael Maire, Serge Belongie, James Hays, Pietro Perona, Deva
  Ramanan, Piotr Doll{\'a}r, and C~Lawrence Zitnick.
\newblock Microsoft coco: Common objects in context.
\newblock In {\em European conference on computer vision}, pages 740--755.
  Springer, 2014.

\bibitem{liu2022robotic}
Wenhai Liu, Weiming Wang, Yang You, Teng Xue, Zhenyu Pan, Jin Qi, and Jie Hu.
\newblock Robotic picking in dense clutter via domain invariant learning from
  synthetic dense cluttered rendering.
\newblock {\em Robotics and Autonomous Systems}, 147:103901, 2022.

\bibitem{neider1993opengl}
Jackie Neider, Tom Davis, and Mason Woo.
\newblock {\em OpenGL programming guide}, volume 478.
\newblock Addison-Wesley Reading, MA, 1993.

\bibitem{rios2013discriminatively}
Reyes Rios-Cabrera and Tinne Tuytelaars.
\newblock Discriminatively trained templates for 3d object detection: A real
  time scalable approach.
\newblock In {\em Proceedings of the IEEE international conference on computer
  vision}, pages 2048--2055, 2013.

\bibitem{shi2021stablepose}
Yifei Shi, Junwen Huang, Xin Xu, Yifan Zhang, and Kai Xu.
\newblock Stablepose: Learning 6d object poses from geometrically stable
  patches.
\newblock In {\em Proceedings of the IEEE/CVF Conference on Computer Vision and
  Pattern Recognition}, pages 15222--15231, 2021.

\bibitem{song2015sun}
Shuran Song, Samuel~P Lichtenberg, and Jianxiong Xiao.
\newblock Sun rgb-d: A rgb-d scene understanding benchmark suite.
\newblock In {\em Proceedings of the IEEE conference on computer vision and
  pattern recognition}, pages 567--576, 2015.

\bibitem{tian2020shape}
Meng Tian, Marcelo~H Ang, and Gim~Hee Lee.
\newblock Shape prior deformation for categorical 6d object pose and size
  estimation.
\newblock In {\em European Conference on Computer Vision}, pages 530--546.
  Springer, 2020.

\bibitem{todorov2012mujoco}
Emanuel Todorov, Tom Erez, and Yuval Tassa.
\newblock Mujoco: A physics engine for model-based control.
\newblock In {\em 2012 IEEE/RSJ International Conference on Intelligent Robots
  and Systems}, pages 5026--5033. IEEE, 2012.

\bibitem{uy2019revisiting}
Mikaela~Angelina Uy, Quang-Hieu Pham, Binh-Son Hua, Thanh Nguyen, and Sai-Kit
  Yeung.
\newblock Revisiting point cloud classification: A new benchmark dataset and
  classification model on real-world data.
\newblock In {\em Proceedings of the IEEE/CVF International Conference on
  Computer Vision}, pages 1588--1597, 2019.

\bibitem{vidal2018method}
Joel Vidal, Chyi-Yeu Lin, Xavier Llad{\'o}, and Robert Mart{\'\i}.
\newblock A method for 6d pose estimation of free-form rigid objects using
  point pair features on range data.
\newblock {\em Sensors}, 18(8):2678, 2018.

\bibitem{wang2019normalized}
He Wang, Srinath Sridhar, Jingwei Huang, Julien Valentin, Shuran Song, and
  Leonidas~J Guibas.
\newblock Normalized object coordinate space for category-level 6d object pose
  and size estimation.
\newblock In {\em Proceedings of the IEEE/CVF Conference on Computer Vision and
  Pattern Recognition}, pages 2642--2651, 2019.

\bibitem{wu2017marrnet}
Jiajun Wu, Yifan Wang, Tianfan Xue, Xingyuan Sun, Bill Freeman, and Josh
  Tenenbaum.
\newblock Marrnet: 3d shape reconstruction via 2.5 d sketches.
\newblock {\em Advances in neural information processing systems}, 30, 2017.

\bibitem{wu2018learning}
Jiajun Wu, Chengkai Zhang, Xiuming Zhang, Zhoutong Zhang, William~T Freeman,
  and Joshua~B Tenenbaum.
\newblock Learning shape priors for single-view 3d completion and
  reconstruction.
\newblock In {\em Proceedings of the European Conference on Computer Vision
  (ECCV)}, pages 646--662, 2018.

\bibitem{xiang2018posecnn}
Yu Xiang, Tanner Schmidt, Venkatraman Narayanan, and Dieter Fox.
\newblock Posecnn: A convolutional neural network for 6d object pose estimation
  in cluttered scenes.
\newblock 2018.

\bibitem{yan2017sim}
Mengyuan Yan, Iuri Frosio, Stephen Tyree, and Jan Kautz.
\newblock Sim-to-real transfer of accurate grasping with eye-in-hand
  observations and continuous control.
\newblock {\em arXiv preprint arXiv:1712.03303}, 2017.

\bibitem{you2021prin}
Yang You, Yujing Lou, Ruoxi Shi, Qi Liu, Yu-Wing Tai, Lizhuang Ma, Weiming
  Wang, and Cewu Lu.
\newblock Prin/sprin: On extracting point-wise rotation invariant features.
\newblock {\em IEEE Transactions on Pattern Analysis and Machine Intelligence},
  2021.

\bibitem{you2022canonical}
Yang You, Zelin Ye, Yujing Lou, Chengkun Li, Yong-Lu Li, Lizhuang Ma, Weiming
  Wang, and Cewu Lu.
\newblock Canonical voting: Towards robust oriented bounding box detection in
  3d scenes.
\newblock In {\em Proceedings of the IEEE/CVF Conference on Computer Vision and
  Pattern Recognition}, 2022.

\end{thebibliography}
}
\appendix
\addcontentsline{toc}{section}{Appendices}

\section*{Supplementary}
\section{Direct 9D Pose Estimation and Segmentation}
\begin{figure}[ht]
    \centering
    \includegraphics[width=0.9\linewidth]{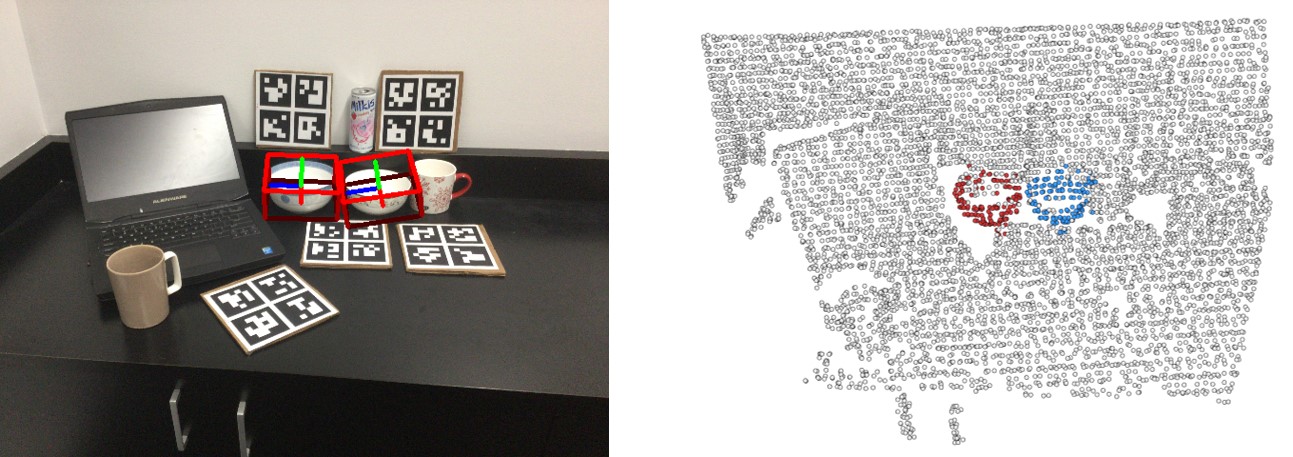}
    \caption{\textbf{Our 9D pose prediction and instance segmentation results on NOCS REAL275 test dataset.} Notice that we do not leverage existing instance segmentation models, and the segmentation is generated by our voting model, which is trained on synthetic objects only.}
    \label{fig:nocsexpnomask}
\end{figure}

In our experiments, we show that our model can work with both segmentation and bounding box masks. Can we locate objects with our voting scheme directly (i.e., without any preprocessing instance detection pipeline)? For some categories like laptop, this is hard since the laptop base can be mixed with the floor in view of point clouds. However, for bowls, this is possible, where the pose estimation is indeed zero-shot without seeing any real-world data during the whole pipeline. We modify Algorithm~1 by sampling point pairs from the whole scene, while keeping the coarse-to-fine procedure. We also use a threshold to generate candidate object locations instead of a simple \textit{argmax} (line 9 in Algorithm~1). 
\paragraph{Zero-shot instance segmentation}
Surprisingly, as a by-product, our method is able to infer the instance-level mask without even seeing any segmentation labels. This is done by counting the contribution of each point, where any point that votes more than $v$ times within a small radius $\epsilon$ of the true object center (line 15 in Algorithm~1), is considered lying on the target object. Qualitative 9D pose and segmentation results are given in Figure~\ref{fig:nocsexpnomask}.  Quantitative results are listed in Table~\ref{tab:zeroshot}.

\section{More Quantitative Results on SUN RGB-D Dataset}
\begin{figure}[ht]
    \centering
    \includegraphics[width=\linewidth]{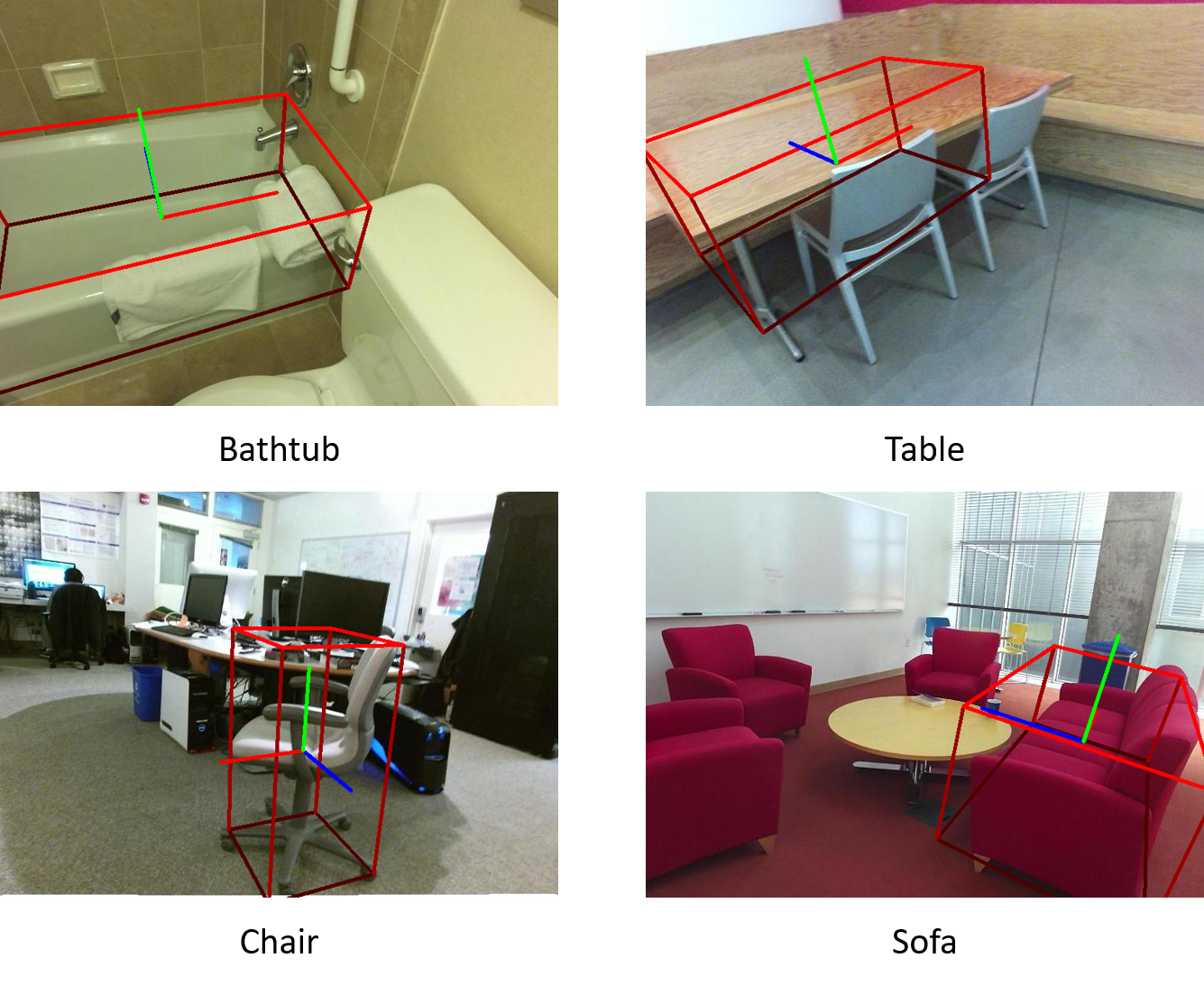}
    \caption{\textbf{Some successful pose estimations on SUN RGB-D.} Notice that our model is trained on synthetic ShapeNet objects only.}
    \label{fig:sunrgbd2}
\end{figure}

We conduct more zero-shot pose estimation experiments on SUN RGB-D, with the instance masks annotated by SUN RGB-D. Notice that our model is trained solely on ShapeNet synthetic models, and then directly tested on SUN RGB-D frames. Results are listed in Table~\ref{tab:moresunrgbd}. Qualitative results are given in Figure~\ref{fig:sunrgbd2}.

\begin{table}[t]
\begin{center}
\small
\begin{tabular}{lccccc}
\toprule
\multirow{3}*{}  & \multicolumn{5}{c}{mAP (\%)}\\
\cmidrule(lr){2-6}
~ & \multirowcell{2}{3D$_{25}$} & \multirowcell{2}{3D$_{50}$} & \multirowcell{2}{5$^\circ$\\ 5 cm} & \multirowcell{2}{10$^\circ$\\ 5 cm} & \multirowcell{2}{15$^\circ$\\ 5 cm} \\
&&&&&\\
\midrule
Bowl & 43.0 & 6.9 & 0.8 & 10.1 & 22.6\\
\bottomrule
\end{tabular}
\end{center}   
\caption{\textbf{Zero-shot 9D pose estimation without detection priors.} We report the mAP results for bowls in real-world scenarios with only synthetic training data.}
\label{tab:zeroshot}
\end{table}

\begin{table}[t]
\begin{center}
\small
\begin{tabular}{lccc}
\toprule
\multirow{3}*{}  & \multicolumn{3}{c}{mAP (\%)}\\
\cmidrule(lr){2-4}
 & \multirowcell{2}{20$^\circ$\\ 10 cm} & \multirowcell{2}{40$^\circ$\\ 20 cm} & \multirowcell{2}{60$^\circ$\\ 30 cm} \\
 &&&\\
\midrule
Bathtub & 10.9 & 38.8 & 49.8 \\
Bookshelf & 0.0 & 1.0 & 6.4 \\
Bed & 0.0 & 0.8 & 3.4 \\
Sofa &  0.0 & 1.7 & 10.5 \\
Table & 0.5 & 6.9 & 17.5 \\
\bottomrule
\end{tabular}
\end{center}
\caption{\textbf{Zero-shot pose estimation results using instance masks provided by SUN RGB-D.} Rotation error is evaluated along the gravity axis.}
\label{tab:moresunrgbd}
\end{table}

\section{More Ablation Studies}
In our sim-to-real pipeline, point clouds are first voxelized and jittered in order to generate the same distribution for both synthetic and real objects. Table~\ref{tab:suppab} gives the ablation studies on these two techniques, where we see that both voxelization and random jittering helps improve the final detection result.

\begin{table}[t]
\begin{center}
\small
\begin{tabular}{lccccc}
\toprule
\multirow{3}*{}  & \multicolumn{5}{c}{mAP (\%)}\\
\cmidrule(lr){2-6}
~ & \multirowcell{2}{3D$_{25}$} & \multirowcell{2}{3D$_{50}$} & \multirowcell{2}{5$^\circ$\\ 5 cm} & \multirowcell{2}{10$^\circ$\\ 5 cm} & \multirowcell{2}{15$^\circ$\\ 5 cm} \\
&&&&&\\
\midrule
No Voxelization  & 76.7 & 21.6 & 11.9 & 37.2 & 45.6\\
No Jittering  & 77.1 & 24.6 & 16.2 & 43.1 & 49.7  \\
\midrule
Ours (full) & \textbf{78.2} & \textbf{26.4} & \textbf{16.9} & \textbf{44.9} & \textbf{50.8}  \\
\bottomrule
\end{tabular}
\end{center}   
\caption{\textbf{Ablation results on NOCS REAL275 test set.}}
\label{tab:suppab}
\end{table}

\section{Detailed results on each category}
\begin{figure}[ht]
    \centering
    \includegraphics[width=\linewidth]{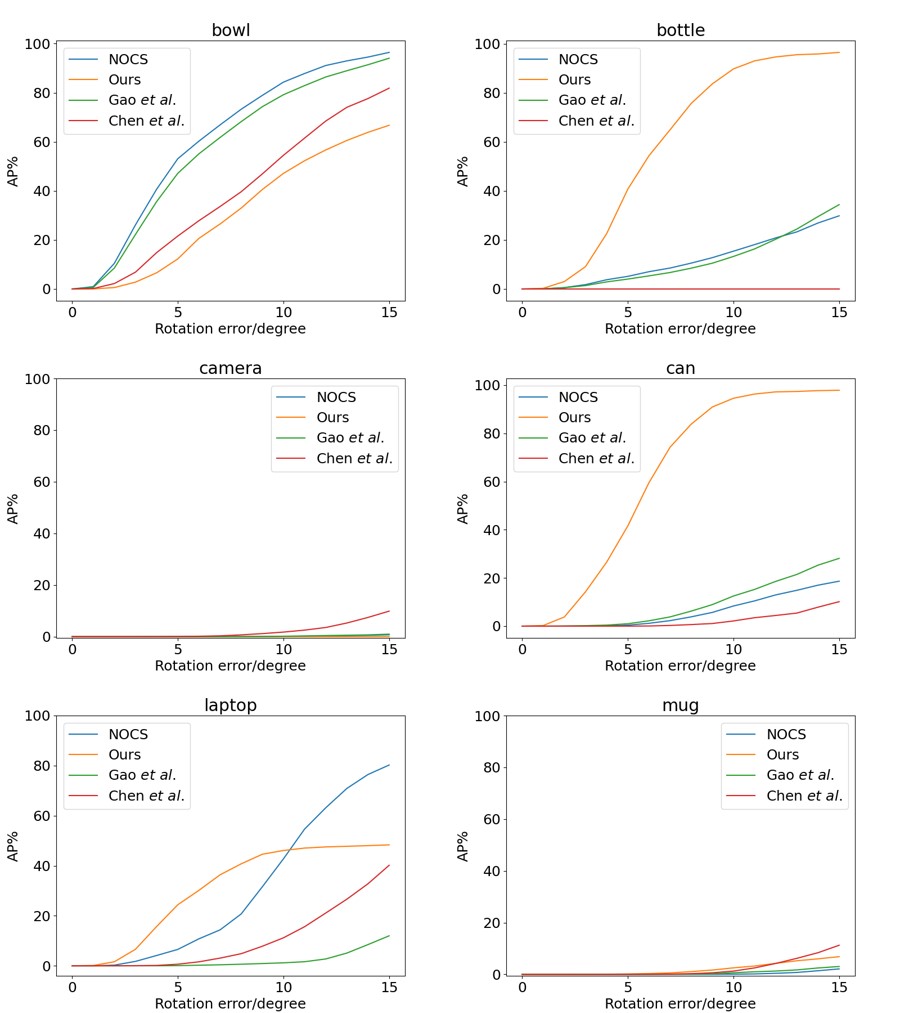}
    \caption{\textbf{Rotation AP for each category on NOCS REAL275 test dataset.}}
    \label{fig:supprot}
\end{figure}

\begin{figure}[ht]
    \centering
    \includegraphics[width=\linewidth]{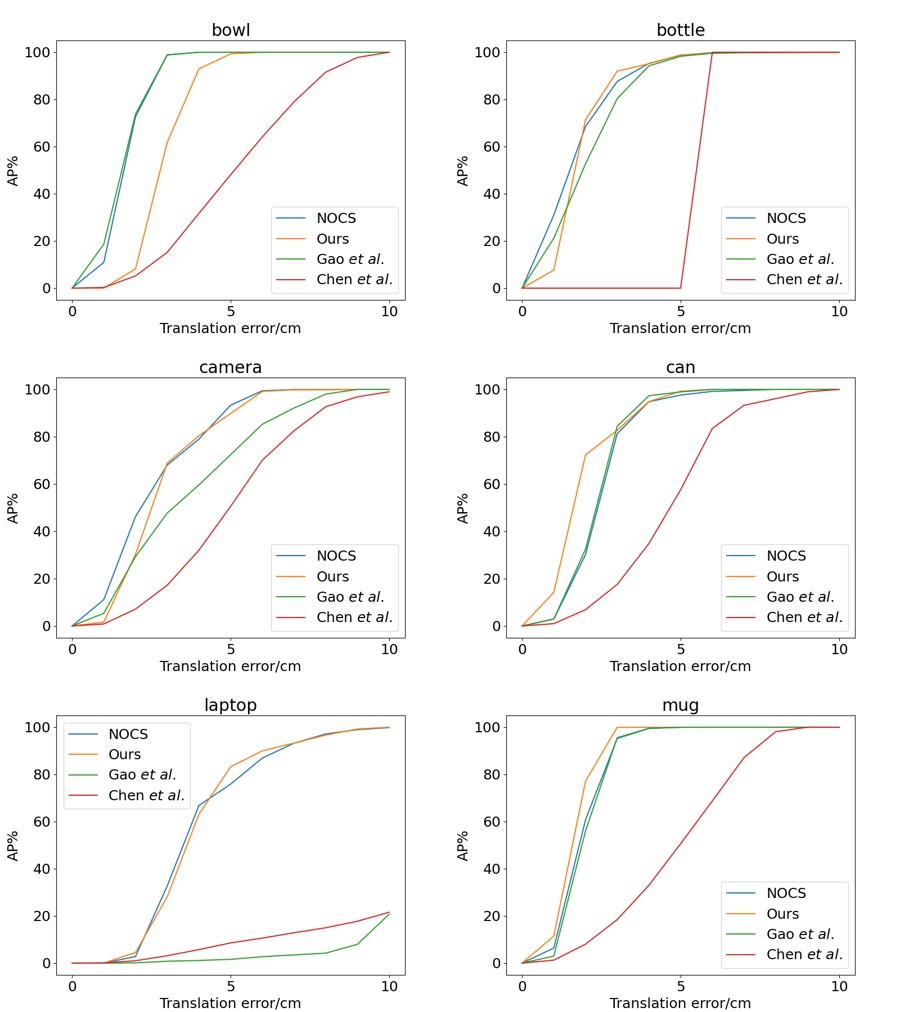}
    \caption{\textbf{Translation AP for each category on NOCS REAL275 test dataset.}}
    \label{fig:supptrans}
\end{figure}

We plot detailed comparisons of each category on NOCS REAL275 test set. Rotation AP is given in Figure~\ref{fig:supprot} and Translation AP is shown in Figure~\ref{fig:supptrans}. Our model achieves the best result on most categories, and outperforms previous self-supervised methods by a large margin.

\end{document}